\documentclass[twoside]{article}

\usepackage[accepted]{aistats2020}
\usepackage[utf8]{inputenc} 
\usepackage[T1]{fontenc}    
\usepackage{url}            
\usepackage{booktabs}       
\usepackage{amsfonts}       
\usepackage{nicefrac}       
\usepackage{microtype}      

\usepackage{enumitem}   

\usepackage{dsfont}
\usepackage{xspace}

\usepackage[dvipsnames,usenames]{xcolor}
\definecolor{NavyBlue}{cmyk}{0.94,0.54,0,0}
\definecolor{DarkBlue}{rgb}{0,0.2,0.6}

\usepackage{hyperref}
\hypersetup{colorlinks=true, linkcolor=BrickRed,citecolor=DarkBlue,urlcolor=DarkBlue}

\usepackage[abbreviations,symbols,nogroupskip,nonumberlist]{glossaries-extra} 

\usepackage{cleveref} 

\usepackage[pdftex]{graphicx}

\usepackage{bm}          
\usepackage{nicefrac}

\usepackage{amssymb,amsmath,amsthm, mathtools}
\usepackage{amscd}
\usepackage{float}    
\usepackage{commath}      

\usepackage{caption, subcaption}

\newcommand{\sref}[1]{\S\ref{#1}}

\usepackage{siunitx} 

\usepackage{xspace}

\usepackage[
    backend=bibtex,
    style=authoryear,
    natbib=true,    
    url=false, 
	isbn=false,
    doi=false,
	maxnames=2, 
	maxbibnames=8, 
	giveninits=true, 
	uniquename=false,
    eprint=true
]{biblatex}
\usepackage{wrapfig}

\usepackage{algorithm}
\usepackage{algorithmic}

\newcommand{\ra}[1]{\renewcommand{\arraystretch}{#1}}

\newtheorem{theorem}{Theorem}[section]

\newtheorem{lemma*}{Lemma}

\def\R{\mathbb{R}}
\def\K{\mathbb{K}}

\def\C{\mathbf{C}}

\def\K{\mathbf{K}}

\def\P{\mathbf{P}}

\def\W{\mathbf{W}}
\def\X{\mathbf{X}}
\def\Y{\mathbf{Y}}

\def\u{\mathbf{u}}
\def\v{\mathbf{v}}

\def\x{\mathbf{x}}
\def\y{\mathbf{y}}

\def\0{\mathbf{0}}

\def\cF{\mathcal{F}}
\def\cM{\mathcal{M}}
\def\cP{\mathcal{P}}

\def\cX{\mathcal{X}}
\def\cY{\mathcal{Y}}

\DeclareMathOperator{\arccosh}{arcosh}

\DeclareMathOperator{\arctanh}{arctanh}

\newcommand{\suchthat}{\;\ifnum\currentgrouptype=16 \middle\fi|\;}

\DeclarePairedDelimiterX{\inner}[2]{\langle}{\rangle}{#1, #2}

\DeclareMathOperator*{\argmin}{argmin}

\DeclarePairedDelimiterX{\infdivx}[2]{(}{)}{%
  #1\;\delimsize\|\;#2%
}

\newenvironment{itemize*}%
  {\begin{itemize}%
  \vspace{-0.5cm}
    \setlength{\itemsep}{0pt}%
    \setlength{\parskip}{0pt}}%
  {\end{itemize}}
\newenvironment{enumerate*}%
{\begin{enumerate}
    \vspace{-0.5cm}
    \setlength{\itemsep}{0pt}%
    \setlength{\parskip}{0pt}}%
  {\end{enumerate}}

  {\list{}{\leftmargin=#1\rightmargin=#1}\item[]}%
  {\endlist}

\newcommand\tabref{Table~\ref}

\newcommand{\En}{\textsc{En}\xspace}
\newcommand{\Es}{\textsc{Es}\xspace}

\newcommand{\Ca}{\textsc{Ca}\xspace}
\newcommand{\Fr}{\textsc{Fr}\xspace}
\newcommand{\It}{\textsc{It}\xspace}

\usepackage{cleveref}
\crefname{section}{§}{§§}
\Crefname{section}{§}{§§}

\runningtitle{Hierarchy Matching with Hyperbolic Optimal Transport}

\begin{document}

\twocolumn[

\aistatstitle{Unsupervised Hierarchy Matching with \\ Optimal Transport over Hyperbolic Spaces}

\aistatsauthor{ David Alvarez-Melis \And Youssef Mroueh \And  Tommi Jaakkola }

\aistatsaddress{ Microsoft Research \And IBM Research \And CSAIL, MIT } ]

\begin{abstract}
   This paper focuses on the problem of unsupervised alignment of hierarchical data such as ontologies or lexical databases. This problem arises across areas, from natural language processing to bioinformatics, and is typically solved by appeal to outside knowledge bases and label-textual similarity. In contrast, we approach the problem from a purely geometric perspective: given only a vector-space representation of the items in the two hierarchies, we seek to infer correspondences across them. Our work derives from and interweaves hyperbolic-space representations for hierarchical data, on one hand, and unsupervised word-alignment methods, on the other. We first provide a set of negative results showing how and why Euclidean methods fail in this hyperbolic setting. We then propose a novel approach based on optimal transport over hyperbolic spaces, and show that it outperforms standard embedding alignment techniques in various experiments on cross-lingual WordNet alignment and ontology matching tasks. 
\end{abstract}
   
\section{Introduction}    
\vspace{-0.2cm}
Hierarchical structures are ubiquitous in various domains, such as natural language processing and bioinformatics.  For example, structured lexical databases like WordNet \citep{miller1995wordnet} are widely used in computational linguistics as an additional resource in various downstream tasks \citep{moldovan2001logic,  shi2005putting, bordes2012joint}. On the other hand, ontologies are often used to store and organize relational data. Building such datasets is expensive and requires expert knowledge, so there is great interest in methods to merge, extend and extrapolate across them. A fundamental ingredient in all of these tasks is \textit{matching}\footnote{Throughout this work, we interchangeably use \textit{matching} and \textit{alignment} to refer to this task.} different datasets, i.e., finding correspondences between their entities. For example, the problem of ontology alignment is an active area of research, with important implications for integrating heterogeneous resources, across domains or languages \citep{spohr2011machine}. We refer the reader to \citep{euzenat2013ontology} for a thorough survey on the state of this problem. On the other hand, there is a long line of work focusing on automatic WordNet construction that seeks to leverage existing large WordNets (e.g., in English) to automatically build WordNets in other low-resource languages \citep{lee2000automatic, saveski2010automatic, pradet2014wonef, khodak2017automated}. 

\citet{euzenat2013ontology} recognize three dimensions to similarity in ontology matching: semantic, syntactic and external. A similar argument can be made for other types of hierarchical structures. Most current methods for aligning such types of data rely on a combination of these three, i.e., in addition to the relations between entities they exploit lexical similarity and external knowledge. For example, automatic WordNet construction methods often rely on access to machine translation systems \citep{pradet2014wonef}, and state-of-the-art ontology matching systems commonly assume large external knowledge bases. Unsurprisingly, these methods perform poorly when no such additional resources are available \citep{shvaiko2013ontology}. Thus, effective fully-unsupervised alignment of hierarchical datasets remains largely an open problem.

Our work builds upon two recent trends in machine learning to derive a new approach to this problem. On one hand, there is mounting theoretical and empirical evidence of the advantage of embedding hierarchical structures in hyperbolic (rather than Euclidean) spaces \citep{nickel2018learning, ganea2018hyperbolic, desa2018representation}. On the other hand, various fully unsupervised geometric approaches have recently shown remarkable success in unsupervised word translation \citep{conneau2018word, Artetxe2018Robust, alvarez-melis2018gromov, grave2019unsupervised, alvarez-melis2019invariances}. We seek to combine these two recent developments by extending the latter to non-Euclidean settings, and using them to find correspondences between datasets by relying solely on their geometric structure, as captured by their hyperbolic-embedded representations. The end goal is a fully unsupervised approach to the problem of hierarchy matching.

In this work, we focus on the second step of this pipeline\;---the \textit{matching}---\;and assume the embeddings of the hierarchies are already learned and fixed. Our approach proceeds by simultaneous registration of the two manifolds and point-wise entity alignment using optimal transport distances. After introducing the building blocks of our approach, we begin our analysis with a set of negative results. We show that state-of-the-art methods for unsupervised (Euclidean) embedding alignment perform very poorly on hyperbolic embeddings, even after modifying them to account for this geometry. This failure is caused by a type of invariance\;---not exhibited by Euclidean embeddings---\;that we refer to as \textit{branch permutation}. At a high level, this phenomenon is characterized by a lack of consistent ordering of branches in the representations of a dataset across different runs of the embedding algorithm (Fig.~\ref{fig:branch_invariance_transported}a), and is akin to the node order invariance in trees.

In response to this challenge, we further generalize our approach by learning a flexible nonlinear registration function between the spaces with a hyperbolic neural network \citep{ganea2018hypernn}. This nonlinear map is complex enough to register one of the hyperbolic spaces (Fig.~\ref{fig:branch_invariance_transported}b), and is learned by minimizing an optimal transport problem over hyperbolic space, which provides both a gradient signal for training and a pointwise (soft) matching between the embedded entities. The resulting method (illustrated in Fig.~\ref{fig:intuition_diagram}) is capable of aligning embeddings in spite of severe branch permutation, which we demonstrate with applications in WordNet translation and biological ontology matching.

In summary, we make the following contributions:
\vspace{-0.2cm}
\begin{itemize}[wide=0pt, leftmargin=\dimexpr\labelwidth + 2\labelsep\relax]
	\itemsep0.5pt
    \item Formulating the problem of unsupervised hierarchy matching from a geometric perspective, casting it as a correspondence problem between hyperbolic spaces
    \item Showing that state-of-the-art methods for unsupervised embedding alignment fail in this task, and find the cause of this to be a unique type of invariance found in popular hyperbolic embeddings
    \item Proposing a novel framework for Riemannian nonlinear registration based on hyperbolic neural networks, which might be of independent interest
    \item Empirically validating this approach with experiments on WordNet hierarchies and ontologies
\end{itemize}

\paragraph{Notation and Conventions}
Let $\cP(\cX)$ be the set of probability distributions over a metric space $\cX$. For a continuous map $f: \cX \rightarrow \cY$ we note by $f^{\sharp}: \cP(\cX) \rightarrow \cP(\cY)$ its associated push-forward operator, i.e., for any $\mu \in \cP(\cX)$,  $\nu = f^{\sharp}(\mu)$ is the push-forward measure satisfying $\int_{\cY}h(y)\dif\nu(y) = \int_{\cX}h(f(x))\dif \mu(x) \medspace \forall h \in \mathcal{C}(\cY)$. The image of $f$ is denoted as  $f[\cX]=\{f(x)\mid x\in \cX\}$. Finally, $\mathrm{O}(n)$ and $\mathrm{SO}(n)$ are the orthogonal and special orthogonal groups of order $n$.

\section{Related Work}

\paragraph{Ontology Matching} Ontology matching is an important problem in various bio-medical applications, e.g., to find correspondences between disease and phenotype ontologies \citep[and references therein]{algergawy2018oaei}. Techniques in ontology matching are usually rule-based, and often rely on entity label similarity and external knowledge bases, making them unfit for unsupervised settings. Here instead we do not assume any additional information nor textual similarity.

\vspace{-0.3cm}
\paragraph{Hyperbolic Embeddings}
Since their introduction \citep{chamberlain2017neural, nickel2017poincare}, research on automatic embedding of hierarchical structures in hyperbolic spaces has gained significant traction \citep{ganea2018hyperbolic, desa2018representation, tay2018hyperbolic}. The main appeal of this approach is that hyperbolic geometry captures several important aspects of hierarchies and other structured data \citep{nickel2017poincare}. We rely on these embeddings to represent the hierarchies of interest.   

\vspace{-0.3cm}
\paragraph{Unsupervised Word Embedding Alignment}
Word translation based on word embeddings has recently gained significant attention after the successful fully-unsupervised approach of \citet{conneau2018word}. Their method finds a mapping between embedding spaces with adversarial training, after which a refinement procedure based on the Procrustes problem produces the final alignment. Various non-adversarial approaches have been proposed since, such as robust-self learning \citep{Artetxe2018Robust}. On the other hand, Optimal transport\;---e.g., Wasserstein---\;distances have been recently shown to provide a robust and effective approach to the problem of unsupervised embedding alignment \citep{zhang2017earth, alvarez-melis2018gromov, grave2019unsupervised, alvarez-melis2019invariances}. For example, \citet{alvarez-melis2019invariances} and \citet{grave2019unsupervised} use a hybrid optimization objective over orthogonal transformations between the spaces (i.e., an Orthogonal Procrustes problem) and Wasserstein couplings between the samples. These works consider only Euclidean settings. Alternatively, this problem can be successfully approached \citep{alvarez-melis2018gromov} with a generalized version of optimal transport, the Gromov-Wasserstein (GW) distance \citep{memoli2011gromov}, which relies on comparing distances between points rather than the points themselves. The recently proposed Fused Gromov-Wasserstein distance \citep{vayer2018fused} extends this to structured domains such as graphs, but as opposed to our approach, assumes node features and knowledge of the full graph structure. While the GW distance provides a stepping stone towards alignment of more general embedding spaces, it cannot account for the type of invariances encountered in practice when operating on hyperbolic embeddings, as we will in Section~\ref{sec:hyperbolic_ot}.

\vspace{-0.3cm}
\paragraph{Correspondence Analysis}
Finding correspondences between shapes is at the heart of many problems in computer graphics. One of the classic approaches to this problem is the Iterative Closest Point method \citep{chen1992object, besl1992method} (and its various generalizations, e.g., by \citet{rusinkiewicz2001efficient}), which alternates between finding (hard) correspondences through nearest-neighbor pairing and finding the best rigid transformation based on those correspondences (i.e., solving a Orthogonal Procrustes problem). The framework we propose can be understood as generalizing ICP in various ways: allowing for Riemannian Manifolds (beyond Euclidean spaces), going beyond rigid (orthogonal) registration and relaxing the problem by allowing for soft correspondences, which the framework of optimal transport naturally provides. 
\section{Hyperbolic Space Embeddings}
\label{sec:background}

A fundamental question when dealing with any type of symbolic data is how to represent it. As the advent of representation learning has proven, finding the right feature representation is as\;---and often more---\;important than the algorithm used on it. Naturally, the goal of such representations is to capture relevant properties of the data. For our problem, this is particularly important. Since our goal is to find correspondences between datasets based purely on their relational structure, it is crucial that the representation capture the semantics of these relations as precisely as possible. 

Traditional representation learning methods embed symbolic objects into low-dimensional Euclidean spaces. These approaches have proven very successful for embedding large-scale co-occurrence statistics, like linguistic corpora for word embeddings \citep{mikolov2013distributed, pennington2014glove}. However, recent work has shown that data for which semantics are given in the form of hierarchical structures is best represented in hyperbolic spaces, i.e., Riemannian manifolds with negative curvature \citep{chamberlain2017neural, nickel2017poincare, ganea2018hyperbolic}. Among the arguments in favor of these spaces is the fact that any tree can be embedded into finite hyperbolic spaces with arbitrary precision \citep{gromov1987hyperbolic}. This stands in stark contrast with Euclidean spaces, for which the dependence on dimension grows exponentially. In practice, this means that very low-dimensional hyperbolic embeddings often perform on-par or above their high-dimensional Euclidean counter parts in various downstream tasks \citep{nickel2017poincare, ganea2018hyperbolic, tay2018hyperbolic}. This too is an appealing argument in our application, as we are interested in matching very large datasets, making computational efficiency crucial.

Working with hyperbolic geometry requires a model to represent it and operate on it. Recent computational approaches have mostly focused on the Poincar\'e Disk\;---or \textit{Ball}, in higher dimensions---\;model. This model is defined by the manifold $\mathbb{D}^d = \{x \in \R^n \suchthat \|x\| < 1\}$, equipped with the metric tensor $g_x^{\mathbb{D}} = \lambda_x^2g^E$, where $\lambda_x:= 1/(1 - \|x\|_2^2)$ is the \textit{conformal factor} and $g^E$ is the Euclidean metric tensor. With this, $(\mathbb{D}^d, g_x^{\mathbb{D}})$ has a Riemannian manifold structure, with the induced Riemannian distance given by:  
\begin{equation}\label{eq:poincare_distance}
    d_{\mathbb{D}}(\u, \v) = \arccosh \left (1 + 2 \frac{\| \u - \v\|^2}{(1\!-\!\|\u\|^2)(1\!-\!\|\v\|^2)} \right),
\end{equation}
which yields the following norm on the Poincar\'e Ball:
\begin{equation}\label{eq:poincare_norm}
    \| \u \|_{\mathbb{D}} = d_{\mathbb{D}}(0, \u)  = 2 \arctanh (\|\u\|).
\end{equation}
It can be seen from this expression that the magnitude of points in the Poincar\'e Ball tends to infinity towards its boundary. This phenomenon intuitively illustrates the tree-like structure of hyperbolic space: starting from the origin, the space becomes increasingly\;---in fact, exponentially more---\;densely packed towards the boundaries, akin to how the width of a tree grows exponentially with its depth.

Hyperbolic embedding methods find representations in the Poincar\'e Ball by constrained optimization (i.e., by imposing $\|\x\|<1$) of a loss function that is often problem-dependent. For datasets in the form of entailment relations $\mathcal{D}= \{(u, v)\}$, where $(u, v) \in \mathcal{D}$ means that $u$ is a subconcept of $v$, \citet{nickel2017poincare} propose to minimize the following soft-ranking loss:
\begin{equation}\label{eq:nickel_loss}
    \mathcal{L}(\Theta) = \sum_{(u,v) \in \mathcal{D}} \log \frac{e^{-d(\u,\v)}}{\sum_{\v'\in \mathcal{N}(u)}e^{-d(\u,\v')}},
\end{equation}
where $\Theta=\{\u\}$ are the embeddings and $\mathcal{N}(u) = \{ v \suchthat (u,v) \not\in \mathcal{D} \}$ is a set of negative examples for $u$.

Transformations in the Poincar\'e Ball will play a prominent role in the development of our approach in Section~\ref{sec:approach}, so we discuss them briefly here. Since the Poincar\'e Ball is bounded, any meaningful operation on it must map $\mathbb{D}^d$ onto itself. Furthermore, for registration we are primarily interested in isometric transformations on the disk, i.e., we seek analogues of Euclidean vector translation, rotation and refection. In this model, translations are given by M\"obius addition, defined as
\begin{equation}\label{eq:mobius_sum}
    \u \oplus\v \triangleq \frac{(1 + 2\langle \u, \v \rangle + \|\v\|_2^2)\u + (1-\|\u\|_2^2)\v}{1+2\langle \u, \v \rangle + \|\u\|_2^2\|\v\|_2^2}.
\end{equation}
This definition conforms to our intuition of translation, e.g., if the origin of the disk is translated to $\v$, then $\x$ is translated to $\v \oplus \x$. Note that this addition is neither commutative nor associative. More generally, it can be shown that all isometries in the Poincar\'e Ball have the form $T(\x) = \P(\v \oplus \x)$, where $\v \in \mathbb{D}^d$ and $\P \in \text{SO}(d)$, i.e., it is an orientation-preserving isometry in $\R^d$. Two other important operations are the logarithmic $\log_{\mathbf{p}}(\cdot)$ and exponential $\exp_{\mathbf{p}}(\cdot)$ maps on a Riemannian manifold, which map between the manifold and its tangent space $T_{\mathbf{p}}\cX$ at a given point $\mathbf{p}$. For the Poincar\'e Ball, these maps can be expressed as
\begin{align*}
    \exp_{\mathbf{p}}(\u) &= \mathbf{p} \oplus \left ( \tanh (\tfrac{1}{2}\lambda_{\mathbf{p}}\|\u\|) \tfrac{\u}{\|\u\|} \right) \\ \log_{\mathbf{p}}(\v) &= \tfrac{2}{\lambda_{\mathbf{p}}}\arctanh \left ( \|(-\mathbf{p}) \oplus \v \| \right) \tfrac{(-\mathbf{p}) \oplus \v}{\|(-\mathbf{p}) \oplus \v \| }.
\end{align*}
    
\section{Matching via Optimal Transport}

\subsection{Optimal Transport Distances}
Optimal transport (OT) distances provide a powerful and principled approach to find correspondences across distributions, shapes, and point clouds \citep{villani2008optimal, Peyre2018Computational}. In its usual formulation, OT considers complete and separable metric spaces $\cX, \cY$, along with probability measures $\alpha \in \cP(\cX)$ and $\beta \in \cP(\cY)$. These can be continuous or discrete measures, the latter often used in practice as empirical approximations of the former whenever working in the finite-sample regime.  \citet{kantorovitch1942translocation} formulated the transportation problem as:
\begin{equation}\label{eq:wasserstein_distance}
    \text{OT}(\alpha, \beta) \triangleq \min_{\pi \in \Pi(\alpha, \beta)} \int_{\cX \times \cY} c(x,y) \dif \pi(x,y)
\end{equation}
where $c(\cdot, \cdot): \cX \times \cY \rightarrow \R^{+}$ is a cost function (the ``ground'' cost), and the set of couplings $\Pi(\alpha, \beta)$ consists of joint probability distributions over the product space $\cX \times \cY$ with marginals $\alpha$ and $\beta$, i.e.,  
\begin{equation}\label{eq:transportation_polytope}
    \Pi(\alpha, \beta) \triangleq \{\pi \in \cP(\cX\times\cY) \suchthat P_{1}^{\sharp}\pi = \alpha,  P_{2}^{\sharp}\pi=\beta \},
\end{equation}
where $P_1(x,y) = x$ and $P_2(x,y)=y$. Whenever $\cX,\cY$ are equipped with a metric $d$, it is natural to use it as ground cost, e.g., $c(x,y) = d(x,y)^p$ for some $p \geq 1$. In such case, $\text{W}_p(\alpha,\beta)\triangleq \text{OT}(\alpha,\beta)^{1/p}$ is called the $p$-Wasserstein distance. The case $p=1$ is also known as the Earth Mover's Distance \citep{rubner2000earth}.

In applications, the measures $\alpha$ and $\beta$ are often unknown, and are accessible only through finite samples $\{\x^{(i)}\} \in \cX, \{\y^{(j)}\} \in \cY$. In that case, these can be taken to be discrete measures $\alpha=\sum_{i=1}^n \mathbf{a}_i\delta_{\x^{(i)}}$ and $\beta=\sum_{i=1}^m \mathbf{b}_i\delta_{\y^{(j)}}$, where $\mathbf{a}$, $\mathbf{b}$ are vectors in the probability simplex, and the pairwise costs can be compactly represented as an $n\times m$ matrix $\C$, i.e., $\C_{ij}=c(\x^{(i)}, \y^{(j)})$. In this case, Equation \eqref{eq:wasserstein_distance} becomes a linear program. Solving this problem scales cubically on the sample sizes, which is often prohibitive in practice. Adding an entropy regularization, namely
\vspace{-0.2cm}
\begin{equation}\label{eq:entreg_wasserstein}
       \text{OT}_{\varepsilon}(\alpha, \beta) \triangleq \min_{\pi \in \Pi(\alpha, \beta)} \!\int_{\cX\!\times\!\cY} \hspace{-10pt}c(x,y) \dif \pi + \varepsilon \text{H}(\pi\!\mid\!\alpha\!\otimes\!\beta)
\end{equation}
where $\text{H}(\pi\!\mid\!\alpha\!\otimes\!\beta)=\int \log(\dif \pi / \dif \alpha \dif \beta)\dif \pi $, leads to a problem that can be solved much more efficiently \citep{altschuler2017near-linear} and which has better sample complexity \citep{genevay2019sample} than the unregularized problem. In the discrete case, Problem~\eqref{eq:entreg_wasserstein} can be solved with the Sinkhorn algorithm \citep{cuturi2013sinkhorn,Peyre2018Computational}, which iteratively updates $\u \gets \mathbf{a} \oslash \K \v$ and $\v \gets \mathbf{b} \oslash \K^{\top}\u$, where $\K\triangleq \exp\{-\frac{1}{\varepsilon}\C\}$ and the division $\oslash$ and exponential are entry-wise. 

\begin{figure*}[t]
    \centering
    \includegraphics[width=0.88\linewidth, trim={0.48cm 0 0 0}, clip]{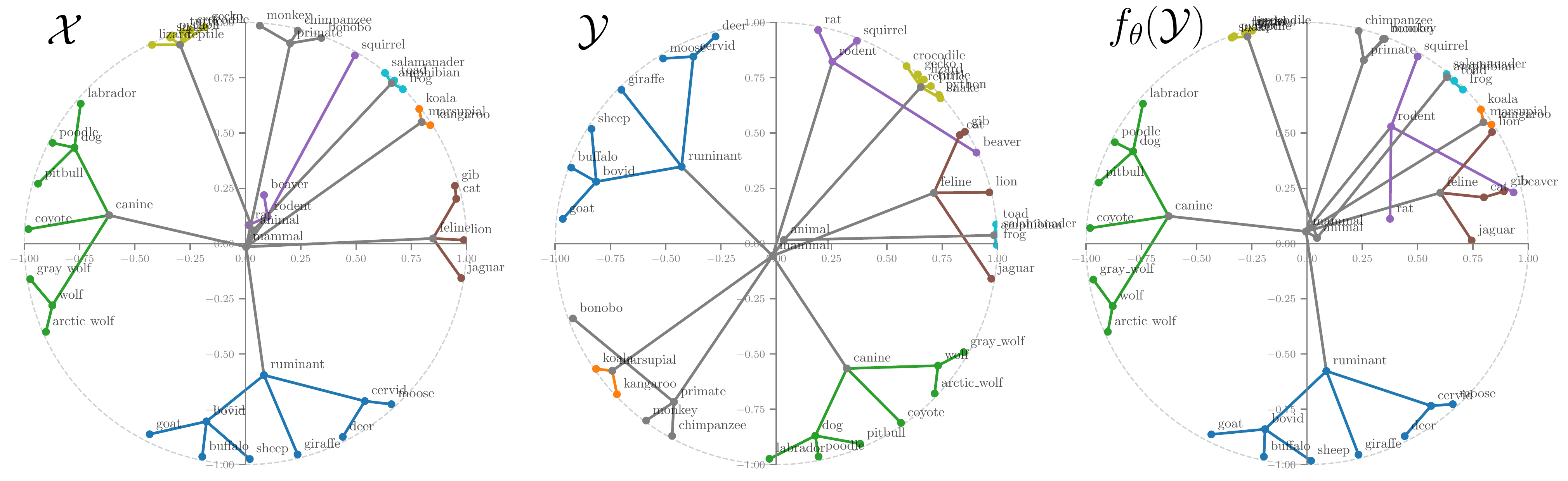}%
    \includegraphics[width=0.13\linewidth]{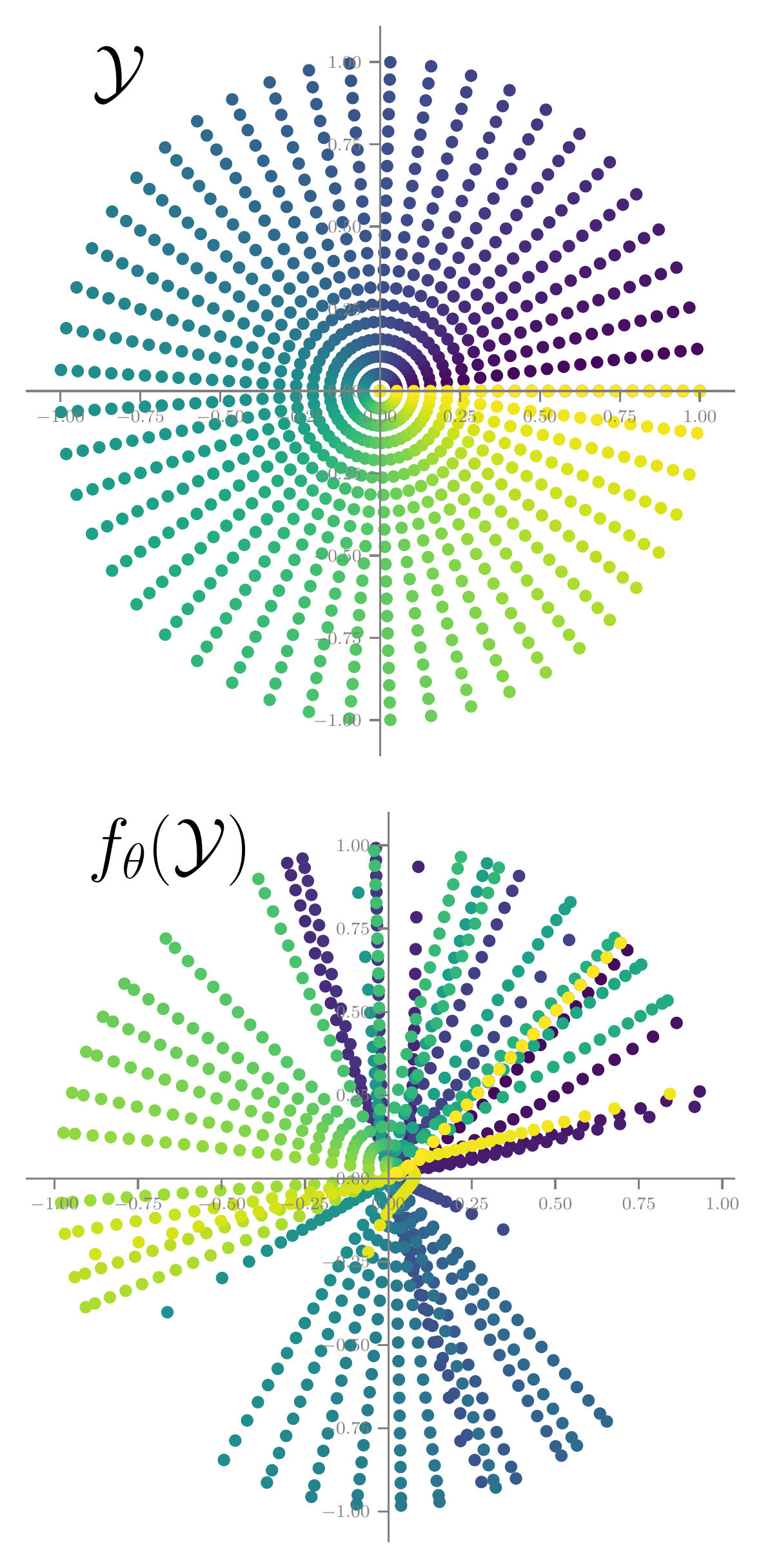}
    \caption{\textbf{Invariance and Registration in the Poincar\'e Disk}. Besides rotation invariance, hyperbolic embeddings can exhibit \textit{branch permutation invariance}, characterized by changes in the relative position of branches (e.g., green and blue here) across embedding instances despite their shape being preserved. The two embeddings $\cX$, $\cY$ shown here were produced by the method of \citet{nickel2018learning} on the same simple hierarchy, using the same hyperparmeters but different random seeds. To allow for unsupervised correspondences between the embeddings to be inferred, the target space must be warped with a registration function $f_{\theta}$ (learnt as part of our method), which is highly non-linear, as shown by the density plots on the right. 
    }
    \label{fig:branch_invariance_transported}
\vskip -0.2in 
\end{figure*}

\subsection{Unsupervised Matching with OT}\label{sec:unsup_matching}
\vspace{-0.2cm}
Besides providing a principled geometric approach to compare distributions, OT has the advantage of producing, as an intrinsic part of its computation, a realization of the optimal way to match the two distributions. Any feasible coupling $\pi \in \Pi(\alpha, \beta)$ in problem \eqref{eq:wasserstein_distance} \;---or \eqref{eq:entreg_wasserstein}---\;can be interpreted as a ``soft'' (multivalued) matching between $\alpha$ and $\beta$. Therefore, the optimal $\pi^*$ corresponds to the minimum-cost way to match them. In the case where the distributions are discrete (e.g., point clouds) $\pi^*$ is a matrix of soft correspondences. Whenever OT is used with the goal of \textit{transportation} (as opposed to just \textit{comparison}), having guarantees on the solution of the problem takes particular importance. Obtaining such guarantees is an active area of research, and a full exposition falls beyond the scope of this work. We provide a brief summary of these in Appendix~\ref{sec:appendix_theory_ot_euclidean}, but refer the interested reader to the survey by \citet{ambrosio2013users}. For our purposes, is suffices to mention that for the quadratic cost (i.e., the 2-Wasserstein distance), the optimal coupling $\pi^*$ is guaranteed to exist, be unique, and correspond to a deterministic map (i.e., a ``hard'' matching).\footnote{Note that $\pi(\alpha, \beta)$ ``includes'' all maps $T:\cX\rightarrow \cY$, which can be expressed as $\pi(\cdot, \cdot) = (Id \times T)^{\sharp}\alpha$.} 

It is tempting to directly apply OT to unsupervised embedding alignment. But note that Problem \eqref{eq:wasserstein_distance} crucially requires that the metric $d$ be congruent across the two spaces (in the Wasserstein case), or more generally, that a meaningful cost function $c$ between them be specified. When the embedding spaces are estimated in a data-driven way, as is usually the case in machine learning, even if these spaces are compatible (e.g., have the same dimensionality) there is no guarantee that the usual metric $d(x,y)$ is meaningful. This could be, for example, because the spaces are defined up to rotations and reflections, creating a class of invariants that the ground metric does not take into account. 

A natural approach to deal with this lack of registration between the two spaces is to simultaneously find a global transformation that corrects for this \textit{and} an optimal coupling that minimizes the transportation cost between the distributions. Formally, in addition to the optimal coupling, we now also seek a mapping $f: \cY \rightarrow \cX$ which realizes 
\begin{equation}\label{eq:invariant_wasserstein}
    \min_{f \in \cF, \pi \in \Pi(\alpha, \beta)} \int_{\cX \times \cY} d(x,f(y)) \dif \pi(x,f(y)).
\end{equation}
where $\cF$ is some function class encoding the invariances. 

As before, we can additionally define an entropy-regularized version of this problem too. Variations of this problem for particular cases of $(\cX,\cY), d(\cdot, \cdot)$ and $\cF$ have been proposed in various contexts, particularly for image registration (e.g., \cite{Rangarajan1997Softassign, cohen1999earth}), and more recently, for word embedding alignment \citep{zhang2017earth, alvarez-melis2019invariances, grave2019unsupervised}. Virtually all these approaches instantiate $\cF$ as the class of orthogonal transformations $\mathrm{O}(d)$, or slightly more general classes of linear mappings \citep{alvarez-melis2019invariances}. In such cases, minimization with respect to $f$ is easy to compute, as it corresponds to an Orthogonal Procrustes problem, which has a closed form solution \citep{gower2004procrustes}. Thus, Problem \eqref{eq:invariant_wasserstein} is commonly solved by alternating minimization. 

\section{Hyperbolic Optimal Transport}\label{sec:hyperbolic_ot}
\vspace{-0.3cm}
In the previous section, we discussed how Wasserstein distances can be used to find correspondences between two embedding spaces in a fully unsupervised manner. However, all the methods we mentioned there have been applied exclusively to Euclidean settings. One might be hopeful that naive application of those approaches on hyperbolic embeddings might just work, but\;---unsurprisingly---\;it does not (cf.~\tabref{tab:wordnet}). Indeed, ignoring the special geometry of these spaces leads to poor alignment. Thus, we now investigate how to adapt such a framework to non-Euclidean settings.

The first fundamental question towards this goal is whether OT extends to more general Riemannian manifolds. The answer is mostly positive. Again, limited space prohibits a nuanced discussion of this matter, but for our purposes it suffices to say that for hyperbolic spaces, under mild regularity assumptions, it can be shown that: (i) OT is well-defined \citep{villani2008optimal}, (ii) its solution is guaranteed to exist, be unique and be induced by a transport map \citep{McCann2001Polar}; and (iii) this map is not guaranteed to be smooth for the usual cost $d_{\mathbb{D}}(x,y)^2$, but it is for variations of it (e.g., $-\cosh \circ d_{\mathbb{D}}$) \citep{lee2012examples}. Further details on why this is the case are provided in Appendix~\ref{sec:appendix_theory_ot_manifold}. These properties justify the use of Wasserstein distances for finding correspondences in the hyperbolic setting of interest. Furthermore, Theorem~\ref{thm:mtw_costs} provides various Riemannian cost functions with strong theoretical foundations and potential for better empirical performance. 

The second step towards generalizing Problem \eqref{eq:invariant_wasserstein} to hyperbolic spaces involves the transformation $f \in \cF$. First, we note that using orthogonal matrices as in the Euclidean case is still valid because, as discussed in Section~\ref{sec:background}, these map the unit disk into itself. Therefore, we can now solve a generalized (hyperbolic) version of the Orthogonal Procrustes problem as before. However, this approach performs surprisingly bad in practice too (see results for \textsc{HyperOT}+Orthogonal $\P$ in \tabref{tab:wordnet}). 

\begin{figure}[t]
    \centering
    \includegraphics[width=\linewidth, trim={0 0.3cm 0 0}, clip]{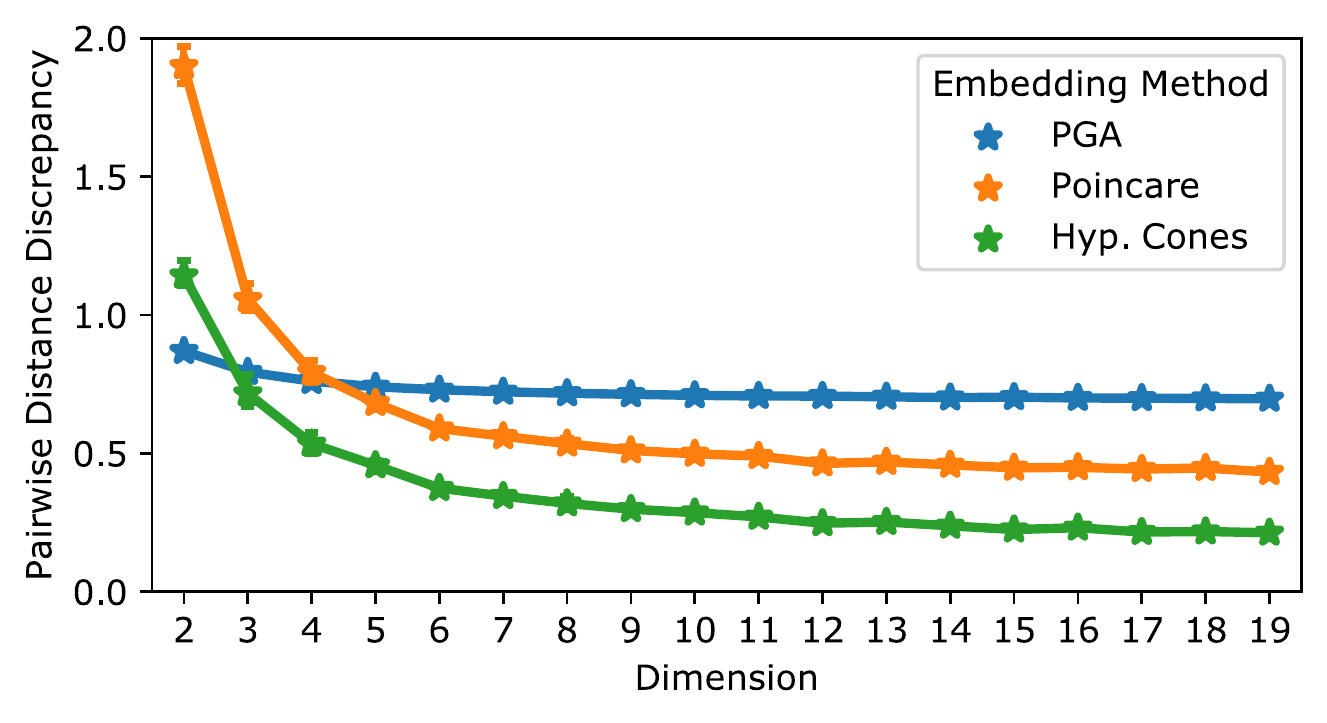}%
    \caption{The branch permutation invariance occurs across hyperbolic embedding methods, even in high dimensions. Here we show the discrepancy (mean $\pm$ 1 s.d.~error bar) over 5 repetitions between pairwise distance matrices of embeddings obtained with different random seeds on the same dataset (\texttt{wordnet-mammals}). Truly isometric embeddings would yield 0 discrepancy.}
    \label{fig:branch_permutation_analysis}
\vskip -0.3in    
\end{figure}

To understand this failure, recall that orthogonality was a natural choice of invariance for embedding spaces that were assumed to differ at most by a rigid transformation, but be otherwise compatible. However, Poincar\'e embeddings exhibit another, more complex, type of invariance, which to the best of our knowledge has not been reported before. It is a \textit{branch permutation invariance}, whereby the relative positions of branches in the hierarchy might change abruptly across different runs of the embedding algorithm, even for the exact same data and hyperparameters. This phenomenon is shown for a simple hierarchy embedded in the Poincar\'e Disk in Figure~\ref{fig:branch_invariance_transported}. While actual discrete trees are indeed invariant to node ordering, \textit{a priori} it is not obvious why this property would be inherited by the embedded space obtained with objective \eqref{eq:nickel_loss}, where non-ancestrally-related nodes do indeed interact (as negative pairs) in the objective. Figure~\ref{fig:branch_permutation_analysis} shows that this phenomenon occurs across hyperbolic embedding methods, such as those of \citet{nickel2017poincare} (Poincar\'e), \citet{ganea2018hyperbolic} (Hyperbolic Cones) and \citet{sala2018representation} (Principal Geodesic Analysis), and that, although less prominent, it is still present in high dimensions.

We conjecture that the cause of this invariance is the use of negative sampling in the embedding objectives, which prioritizes distance preservation between entities that are ancestrally related in the hierarchy, at the cost of downweighting distances between unrelated entities.  A formal explanation of this phenomenon is left for future work. Here, instead, we develop a framework to account and correct these invariances while simultaneously aligning the two embeddings. 

\section{A Framework for Correspondence across Hyperbolic Spaces}

\label{sec:approach}

\begin{figure*}[t]
    \centering
    \includegraphics[width=\linewidth, trim={0 21cm 0 0}, clip]{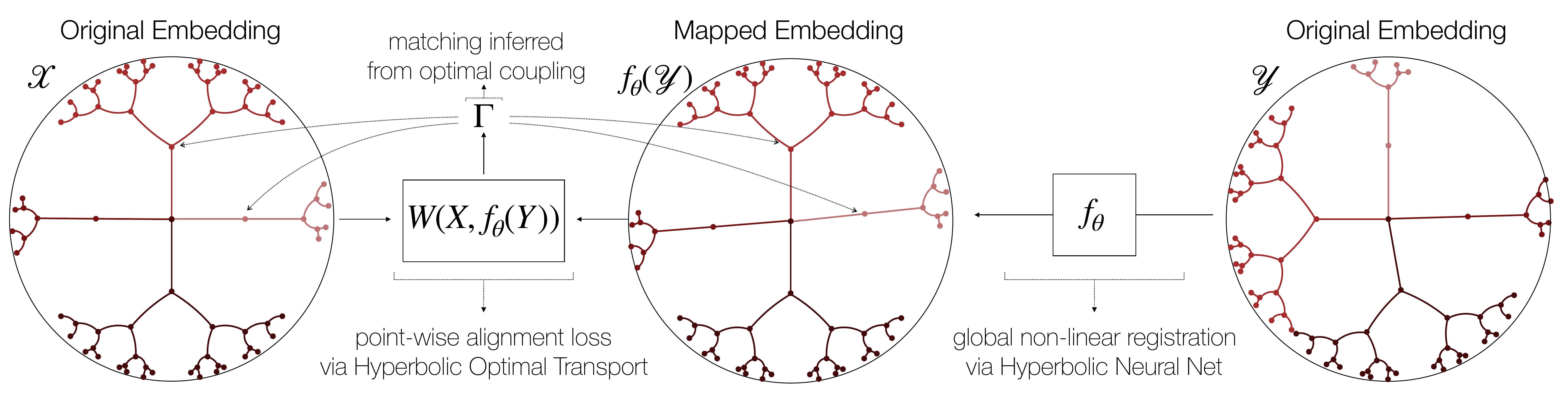}%
    \caption{Schematic representation of the proposed approach. A deep network $f_{\theta}$ globally registers the two hyperbolic embedding spaces ($\cX$ and $\cY$) by correcting for non-linear branch permutations, so that source and mapped target points can be aligned using a hyperbolic variant of Wasserstein distance. Training is done end-to-end in a fully unsupervised way -- no prior known correspondences between the hierarchies are assumed.}
    \label{fig:intuition_diagram}
\end{figure*}

The failure of the baseline Euclidean alignment methods (and their hyperbolic versions), combined with the underlying branch permutability invariance responsible for it, make it clear that the space of registration transformations $\cF$ in Problem~\eqref{eq:invariant_wasserstein} has to be generalized not only beyond orthogonality but beyond linearity too. 

Ideally we would search among all continuous mappings between $\cY$ and $\cX$, i.e, taking $\cF=\{f:\cY \rightarrow \cX \suchthat f\in \mathcal{C}(\cY)\}$. To make this search computationally tractable, we can instead approximate this function class with deep neural networks $f_\theta$ parametrized by $\theta \in \Theta$. 
While an alternating minimization approach is still possible, solving for $\theta$ to completion in each iteration is undesirable. Instead, we reverse the order of optimization and rewrite our objective $  \min_{f_{\theta} \in \cF}  \text{W}_{\varepsilon}(\alpha, f_{\theta}^{\sharp}\beta)$ as
\begin{equation*}
     \min_{f_{\theta},  \pi } \int_{\cX\!\times\!\cY}  d(x,f_{\theta}(y)) \dif \pi(x,f_{\theta}(y))  + \varepsilon \text{H}(\pi\!\mid\!\alpha\!\otimes\!\beta).    
\end{equation*}
\vskip -0.15 in
Since  $\text{W}_{\varepsilon}(\alpha, f_{\theta}^{\sharp}\beta)$ is differentiable with respect to $\theta$, we can use gradient-based methods to optimize it. Wasserstein distances have been used as loss functions before, particularly in deep generative modeling \citep{Arjovsky2017Wasserstein, genevay2018learning,salimans2018improving}. When used as a loss function, the entropy-regularized version (Eq.~\eqref{eq:entreg_wasserstein}) has the undesirable property that $\text{W}_{\varepsilon}(\alpha, \alpha) \neq 0$, in addition to having biased sample gradients \citep{bellemare2017cramer}. Following \citet{genevay2018learning}, we instead consider the \textit{Sinkhorn Divergence}:
\begin{equation}\label{eq:sinkhorn_divergence}
    \text{SD}_{\varepsilon}(\alpha, \beta) \triangleq \text{W}_{\varepsilon}(\alpha,\beta) - \nicefrac{1}{2}\bigl( \text{W}_\varepsilon(\alpha, \alpha)\!+\! \text{W}_{\varepsilon}(\beta,\beta) \bigr).
\end{equation}
Besides being a true divergence and providing unbiased gradients, this function is convex, smooth and positive-definite \citep{feydy2019interpolating}, and its sample complexity is well characterized \citep{genevay2019sample}, all of which make it an appealing loss function. Using this divergence in place of the Wasserstein distance above yields our final objective:
\begin{equation}\label{eq:objective}
    \min_{\theta: f_{\theta}[\mathbb{D}^d] \subseteq \mathbb{D}^d} \text{SD}_{\varepsilon}(\alpha, f_{\theta}^{\sharp}\beta).
\end{equation}
\vskip -0.1 in

The last remaining challenge is that we need to construct a class of neural networks that parametrizes $\cF := \{f_\theta \suchthat f_\theta(\mathbb{D}^d) \subseteq \mathbb{D}^d\}$, i.e., functions that map $\mathbb{D}^d$ onto itself. In recent work, \citet{ganea2018hyperbolic} propose a class of hyperbolic neural networks that do precisely this. As they point out, the basic operations in hyperbolic space that we introduced in Section~\ref{sec:background} suffice to define analogues of various differentiable building blocks of traditional neural networks. For example, a hyperbolic linear layer can be defined as 
$$f_{\textsc{HypL}}(\x; \W, \mathbf{b})\triangleq  (\W\otimes \x) \oplus \mathbf{b} = \exp_0(\W\log_0(\x)) \oplus \mathbf{b}.$$
Analogously, applying a nonlinearity $\sigma(\cdot)$ in the hyperbolic sense can be defined as $\sigma_{\mathbf{D}}(\x) \triangleq \exp_0(\sigma\log_0(\x))$. Here, we also consider \textit{M\"obius Transformation layers}, $f_{\text{M\"obius}}(\x) = \P(\v \oplus \x)$, with $\P \in \text{SO}(d)$ and $\v \in \mathbb{D}^d$. With these building blocks, we can parametrize highly nonlinear maps $f_{\theta}: \mathbb{D}^n \rightarrow \mathbb{D}^n$ as a sequence of such hyperbolic layers, i.e., $\mathbf{h}^{(i)} = \sigma_{\mathbb{D}} ( \W^{(i)}\otimes \mathbf{h}^{(i-1)} \oplus \mathbf{b}^{(i)})$. Note that the hyperbolic linear\;---but \textit{not} the M\"obius---\; layers allow for intermediate vectors $\mathbf{h}^{(i)}$ that are of different size as the input and output, i.e., we can use rectangular weight matrices $\W$ to map intermediate states to Poincar\'e balls of different dimensionality. 

The overall approach is summarized in Figure \ref{fig:intuition_diagram}.

\vspace{-0.3cm}
\paragraph{Optimization}
Evaluation of the loss function in \eqref{eq:objective} is itself an optimization problem, i.e., solving instances of regularized optimal transport. We backpropagate through this objective as proposed by \citet{genevay2018learning}, using the \texttt{geomloss} toolbox for efficiency. For the outer-level optimization, we rely on Riemannian gradient descent \citep{zhang2016riemannian, wilson2018gradient}. We found that the adaptive methods of \citet{becigneul2019riemannian} worked best, particularly \textsc{Radam}. Note that for the \textsc{HyperLinear} layers only the bias term is constrained (on the Poincar\'e Ball), while for our \textsc{M\"obius} layers the weight matrix is also constrained (in the Stiefel manifold), hence for these we optimize over the product of these two manifolds. Additional details are provided in the Appendix.

\vspace{-0.3cm}
\paragraph{Avoiding poor local minima}
The loss function \eqref{eq:objective} is highly non-convex with respect to $\theta$, a consequence of both the objective itself and the nature of hyperbolic neural networks \citep{ganea2018hyperbolic}. As a result, we found that initialization plays a crucial role in this problem, since it is very hard to overcome a poor initial local minimum. Even suitable layer-wise random initialization of weights and biases proved futile. As a solution, we experimented with three pre-training initialization schemes, that roughly ensure (in different ways) that $f$ does not initially ``collapse'' the space $\cY$ (details provided in Appendix~\ref{sec:pretrain}). In addition, we use an annealing scheme on the entropy-regularization parameter $\varepsilon$ \citep{kosowsky1994invisible,alvarez-melis2019invariances}. Starting from an aggressive regularization (large $\varepsilon_0$), we gradually decrease it with a fixed decay rate $\varepsilon_t = \xi \cdot \varepsilon_{t-1}$. We use $\xi = 0.99$ in all experiments. 


\begin{table*}[t]
  \footnotesize
  \centering
  \ra{1.27}
    \begin{subtable}[t]{.3\linewidth}
      \flushleft
		\begin{tabular}{r p{0.5cm} p{0.5cm}}
			\toprule
			 & \textsc{P@1} & P@10 \\
			\midrule
			 \textsc{Full model} \hspace{0.3cm}	& 22.2 & 88.8 \\
			 small 					& 8.7 & 38.0 \\
			 euclidean  				& 3.1 & 13.0 \\
			 elu$\rightarrow$relu  & 6.9 & 37.6 \\
			 \textsc{radam}$\rightarrow$\textsc{rsgd}  & 14.7  &  69.5 \\
			 \textsc{M\"obius} layers 		& 11.9 & 54.3 \\
			 cost: $-\cosh \circ d$ 		& 16.6 & 70.2 \\
			 no pretrain 		& 0.1 & 0.2 \\
			\bottomrule
		\end{tabular}
        \caption{Ablation on \textsc{En}$\rightarrow$\En.}\label{tab:ablation}
    \end{subtable}%
    \begin{subtable}[t]{.7\linewidth}
      \flushright
   \ra{1.2}
  \begin{tabular}{@{}r p{0.5cm} c c c c c c c c@{}}\toprule
 	   & & \multicolumn{2}{c}{\En-\Es} & \multicolumn{2}{c}{\En-\It}  & \multicolumn{2}{c}{\En-\Fr}  & \multicolumn{2}{c}{\En-\Ca}  \\
	   \cmidrule(lr){3-4} \cmidrule(lr){5-6}  \cmidrule(lr){7-8}  \cmidrule(lr){9-10}
	    & T(m) & $\rightarrow$ & $\leftarrow$ & $\rightarrow$ & $\leftarrow$ & $\rightarrow$ & $\leftarrow$ & $\rightarrow$ & $\leftarrow$   \\
	    \midrule
	    \textit{Baseline Euclidean} \\
		\textsc{Muse} & 42 & 0.60 & 1.20 & 0.06 & 0.45 & 0.06 & 2.22 & 0.09 & 0.87 \\
		\textsc{Self-Learn}  & 3 & 0.31 & 1.40 & 0.46 & 0.46 & 0.55 & 0.40 & 0.25 & 0.34 \\
		\textsc{Invar-OT}   & 4 & 0.25 & 0.25 & 2.15 & 0.60 & 0.38 & 2.14 & 0.51 & 7.65\\
	    \textit{Hyperbolic Methods}  \\
		\textsc{Hyp-OT} +  Orth.~$P$& 13 & 4.51 & 5.29 & 11.2 &  0.47 & 5.32 & 4.68 & 7.21 & 5.55   \\		
		\textsc{Hyp-OT} + NN $f_{\theta}$ & 21 & 43.9 & 56.8  & 48.0 & 60.1 & 54.3 & 57.5 & 38.4 & 57.4  \\		
    \bottomrule
  \end{tabular}%
     \caption{P@10 for pair-wise language matching. T(m): runtime in minutes.}\label{tab:wordnet}
    \end{subtable} 
    \caption{Results on the mutlilingual wordnet matching task.}	
\end{table*}

\vspace{-0.3cm}
\section{Experiments}
\vspace{-0.3cm}
\paragraph{Datasets}
For our first set of experiments, we extract subsets of WordNet \citep{miller1995wordnet} in five languages. For this, we consider only nouns and compute their transitive closure according to hypernym relations. Then, for each collection we generate embeddings in the Poincar\'e Ball of dimension 10 using the method of \citet{nickel2018learning} with default parameters.  We release the multi-lingual WordNet dataset along with our codebase. In Section~\ref{sec:noise_sensitivity}, we perform synthetic experiments on the \texttt{cs-phd} network dataset \citep{rossi2015network} again embedded with same algorithm. In addition, we consider two subtasks of the OAEI 2018 ontology matching challenge \citep{algergawy2018oaei}: \textsc{Anatomy}, which consists of two ontologies; and \textsc{biodiv}, consisting of four. Additional details on all the datasets are provided in Table~\ref{tab:dataset_details} in the Appendix.

\vspace{-0.3cm}
\paragraph{Methods}
We first compare ablated versions of our \textsc{Hyperbolic-OT} model (Table~\ref{tab:ablation_details} in the Appendix), and then we compare against three state-of-the-art unsupervised word embedding alignment models: \textsc{Muse} \citep{conneau2018word}, \textsc{Self-Learn} \citep{Artetxe2018Robust} and \textsc{InvarOt} \citep{alvarez-melis2019invariances}, using their recommended configurations and settings. 

\vspace{-0.3cm}
\paragraph{Metrics}
All the baseline methods return transformed embeddings. Using these, we retrieve nearest neighbors and report precision-at-k, i.e., $\text{P}@k=\alpha$ if the true match is within the top $k$ retrieved candidate matches for $\alpha$ percent of the test examples.

\subsection{Multilingual WordNet Alignment}
We first perform an ablation study on the various components of our model in a controlled setting, where the correspondences between the two datasets are perfect and unambiguous. For this, we embed the same hierarchy (the \textsc{En} part of our WordNet dataset) twice, using the same algorithm with the same hyperparameters, but different random seeds. We then evaluate the extent to which our method can recover the correspondences. Starting from our \textsc{Full Model}, we remove and/or replace various components and re-evaluate performance. The exact configuration of the ablated models is provided in Appendix \ref{sec:configs}. The results in \tabref{tab:ablation} suggest that the most crucial components are the use of the appropriate Poincar\'e metric and the pretraining step.
Next, we move on to the real task of interest: matching WordNet embeddings across different languages. Naturally, in this case there might not be perfect correspondences across the entities in different languages. As before, we report Precision@10 and compare against baseline models in \tabref{tab:wordnet}.

\subsection{Noise Sensitivity}\label{sec:noise_sensitivity}
\begin{figure}[t]
    \centering
    \includegraphics[height=3.3cm, trim={0 0 0 0}, clip]{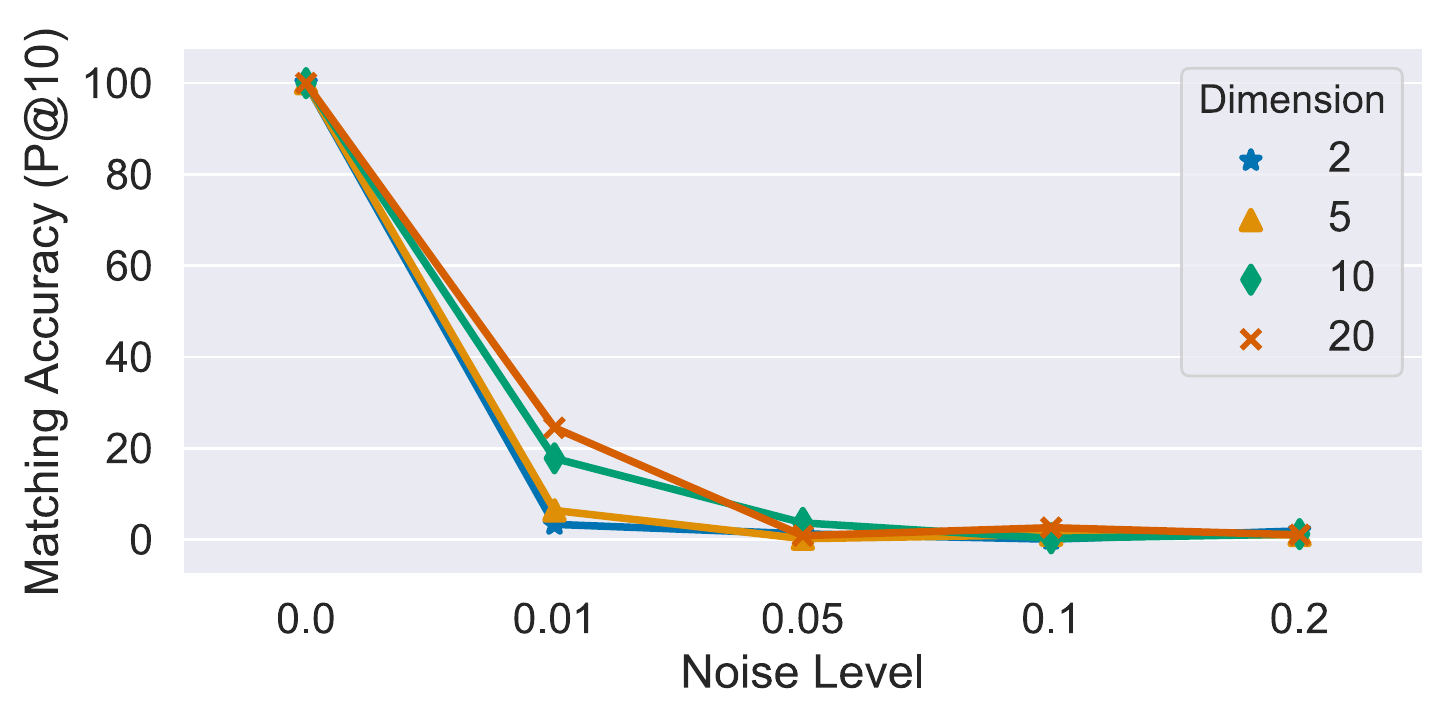}%
    \caption{Noise sensitivity.}
    \label{fig:noise_sensitivity}
\end{figure}

We next analyze the effect of domain discrepancy on the matching quality. Using our method to match a noise-less and noisy versions of the \texttt{csphd} dataset (details in the Appendix), we observe that accuracy degrades rapidly with noise, although ---as expected--- less so for higher-dimensional embeddings (Fig.~\ref{fig:noise_sensitivity}).

\subsection{Ontology Matching}
Finally, we test our method on the OAEI tasks. The results (\tabref{tab:oaei}) show that our method again decidedly outperforms the baseline Euclidean methods, but now the overall performance of all methods is remarkably lower, which suggests the domains' geometry is less coherent (partly because of their vastly different sizes) and/or the correspondences between them more noisy.


\begin{table}[H]
  \scriptsize
  \centering
  \ra{1.3}
  \begin{tabular}{r c c c c c c}
			\toprule
			& \multicolumn{2}{c}{Anatomy} & \multicolumn{4}{c}{Biodiv}  \\
			\cmidrule(lr){2-3} \cmidrule(lr){4-7}
			& H$\rightarrow$M  &  M$\rightarrow$H & F$\rightarrow$P & P$\rightarrow$F & E$\rightarrow$S & S$\rightarrow$E \\
			\midrule 
			\textsc{Muse}					& 0.12 & 0.00 & 3.23 & 0.00 & 0.00 & 0.00 \\
			\textsc{Self-Learn}					& 0.00 & 0.00 & 4.00 & 0.00 & 0.01 & 0.02 \\
			\textsc{Hyp-OT}					& 7.89 & 4.49 & 16.67 & 8.73 & 6.25 & 9.66 \\
			\bottomrule
		\end{tabular}
\caption{Results on the OAEI ontology matching tasks.}\label{tab:oaei}
\end{table} 

\section{Discussion and Extensions}

The framework for hierarchical structure matching proposed here admits various extensions, some of them immediate. We focused on the particular case of the Poincar\'e Ball, but since most of the components of our approach\;---optimization, registration, optimal transport---\;generalize to other Riemannian manifolds, our framework would too. As long as optimizing over a given manifold is tractable, our framework would enable computing correspondences across instances of it. On the other hand, we purposely adopted the challenging setting where no additional information is assumed. This setting is relevant both for extreme practical cases and to stress-test the limits of unsupervised learning in this context. However, our method would likely benefit from incorporating any additional available information as state-of-the-art methods for ontology matching do. In our framework, this information could for example be injected intro the transport cost objective.

\clearpage
\pagebreak

\section*{Acknowledgements}
The work was partially supported by the MIT-IBM project on adversarial learning of multi-modal and structured data.

\printbibliography

\pagebreak
\clearpage

\appendix
\section{Pretraining Strategies}\label{sec:pretrain}

The loss function \eqref{eq:objective} is highly non-convex with respect to $\theta$, a consequence of both the objective itself and the nature of hyperbolic neural networks \citep{ganea2018hyperbolic}. As a result, we found that initialization plays a crucial role in this problem, since it is very hard to overcome a poor initial local minimum. Even layer-wise random initialization of weights and biases proved futile. As a solution, we experimented with the following three pre-training initialization schemes, all of which intuitively try to approximately ensure (in different ways) that $f$ does not ``collapse'' the space $\cY$:

\vspace{-0.3cm}
\paragraph{\textsc{Identity}.} Initialize $f_{\theta}$ to approximate the identity: $$\min_{\theta} \sum_{i=1}^n d_{\mathbb{D}}(\y_i, f_{\theta}(\y_i)),$$ which trivially ensures that $f_{\theta}$ (approximately) preserves the overall geometry of the space.

\vspace{-0.3cm}
\paragraph{\textsc{CrossMap}.} Initialize $f_{\theta}$ to approximately match the target points to the source points in a random permuted order:  $$\min_{\theta} \sum_{i=1}^n d_{\mathbb{D}}(\x_{\sigma(i)}, f_{\theta}(\y_i))$$ for some permutation $\sigma({i})$, which again ensures that $f_{\theta}$ approximately preserves the global geometry, albeit for an arbitrary labeling of the points.

\vspace{-0.3cm}
\paragraph{\textsc{Procrustes}.} Following \citep{bunne2019learning}, we initialize $f_{\theta}$ to be approximately end-to-end orthogonal: $$\min_{\theta} \sum_{i=1}^n d_{\mathbb{D}}(f(\y_i), \P\y_i),$$ where $\P=\argmin_{\P \in \mathrm{O}(n)}{\|\X -\P\Y\|_2^2}$, i.e., $\mathbf{P}$ is the solution of (a hyperbolic version of) the Orthogonal Procrustes problem for mapping $\Y$ to $\X$, which can be obtained via singular value decomposition (SVD). This strategy thus requires computing an SVD for every gradient update on $\theta$; hence, it is significantly more computationally expensive than the other two.

\section{Optimization Details}\label{sec:opt_details}

Each forward pass of the loss function \eqref{eq:objective} requires solving three regularized OT problems. While this can be done to completion in $O(N^2 \log N\varepsilon^{-3})$ time \citep{altschuler2017near-linear}, practical implementations often run the Sinkhorn algorithm for a fixed number of iterations with a tolerance threshold on the objective improvement. We rely on the \texttt{geomloss}\footnote{\url{https://www.kernel-operations.io/geomloss/}} package for efficient differentiable Sinkhorn divergence implementation and on the \texttt{geoopt}\footnote{\url{https://geoopt.readthedocs.io/en/latest/}} package for Riemannian optimization. We run our method for a fixed number of outer iterations (200 in all our experiments), which given the decay strategy on the entropy regularization parameter $\varepsilon$, ensures that $\varepsilon$ ranges from \num{1e1} to \num{1e-2}. All experiments where run a single machine with 32-core processor, Intel Xeon CPU @3.20 GHz, and exploiting computations on the GPU (a single GeForce Titan X) whenever possible. With this configuration the total runtime of our method on the experiments ranged from $<1$ to $~20$ minutes.

\begin{table*}[t]
  \scriptsize
  \centering
  \ra{1.28}
      \flushleft
         \resizebox{\linewidth}{!}{%
		\begin{tabular}{r  c c c c c c c c c c} 
			\toprule
			& \multicolumn{4}{c}{WordNet} & \multicolumn{2}{c}{Anatomy}  & \multicolumn{4}{c}{Biodiv}  \\
	          \cmidrule(lr){2-5} \cmidrule(lr){6-7}  \cmidrule(lr){8-11}
			  & English (\En) & Spanish (\Es)  & French (\Fr) & Catalan (\Ca) & Human & Mouse & \textsc{Flopo} & \textsc{Pto} & \textsc{Envo} &  \textsc{Sweet} \\
			\midrule
			 Entities  & 8206 & 8206 & 8206 & 8206 & 3298 & 2737 & 360 & 1456 & 6461 & 4365 \\
			 Relations & 47938 & 47938 & 47938 & 47938 & 18556 & 7364 & 472 & 11283 & 73881 & 30101 \\
			 Embedding Size  & 10 & 10 & 10 & 10 & 10 & 10 & 10 & 10 & 10 & 10 \\
			\bottomrule
		\end{tabular}%
    }
    \caption{Dataset characteristics. All datasets embedded with the method of \citet{nickel2017poincare}. }\label{tab:dataset_details}
\end{table*}%

\begin{table*}[ht]
  \scriptsize
  \centering
  \ra{1.28}
      \flushleft
         \resizebox{\linewidth}{!}{%
		\begin{tabular}{r  c c c c c c c c c c} 
			\toprule
			 Model & Metric & Cost & Pretrain & $\varepsilon$-annealing & Layers & Hidden dim & Layer Type & Nonlin. & Opt & LR  \\
			\midrule
			 \textsc{Full} & Poincare & $d(x,y)$ & \textsc{CrossMap} & $\num{e1} \rightarrow \num{e-2}$ & 10 & 20 & \textsc{HyperLinear} & elu & \textsc{Radam} & \num{e-3} \\
			 \textsc{Small} & -- & -- & -- & -- & 2 & 10 & -- & -- & -- & -- \\
			 \textsc{Euclidean} & Euclidean & -- & -- & -- & -- & -- & -- & -- & -- & -- \\
			 \textsc{ReLU} & -- & -- & -- & -- & -- & -- & \textsc{ReLU} & -- & -- \\
			 \textsc{Rsgd} & -- & -- & -- & -- & -- & -- & -- & -- & \textsc{Rsgd} & \num{5e-2} \\
			 \textsc{M\"obius} & -- & -- & -- & -- & -- & 10 & \textsc{M\"obius} & -- & -- & -- \\
			 \textsc{$\cosh$ Cost} & -- & $-\cosh \circ d_{\mathbb{D}}$ & -- & -- & -- & -- & -- & -- & -- & --  \\
			 \textsc{No Pretrain}  & -- & -- & None & -- & -- & -- & -- & -- & -- & -- \\
			\bottomrule
		\end{tabular}%
    }
    \caption{Ablated model configurations for the monolingual \En$\rightarrow$\En WordNet task.}\label{tab:ablation_details}
\end{table*}%

\section{Dataset Details}\label{sec:dataset_details}
To generate the parallel WordNet datasets, we use the \texttt{nltk} interface to WordNet, and proceed as follows. In the English WordNet, we first filter out all words except nouns, and generate their transitive closure. For each of the remaining synsets, we query for lemmas in each of the four other languages (\Es, \Fr, \It, \Ca), for which \texttt{nltk} provides multilingual support in WordNet. These tuples of lemmas form our ground-truth translations, which are eventually split into a validation set of size 5000, leaving all the other pairs for test data (approximately 1500 for each language pairs). Note that the validation is for visualization purposes only, and all model selection is done in a purely unsupervised way based on the training objective. After the multi-lingual synset vocabularies have been extracted, we ensure their transitive closures are complete and write all the relations in these closures to a file, which will be used as an input to the \textsc{PoincareEmbeddings} toolkit.\footnote{\url{https://github.com/facebookresearch/poincare-embeddings}}

To generate the datasets for the synthetic noise-sensitivity experiments (\sref{sec:noise_sensitivity}), we start from the original CS-PhD dataset.\footnote{\url{http://networkrepository.com/CSphd.php}} Given a pre-defined value $\nu$, we iterate through the hierarchy removing node $x$ with probability $p$, connecting $x$'s children with $x$'s parent to keep the tree connected. We repeat this with noise values $p \in \mathfrak{P} = [0.01, 0.05, 0.1, 0.2]$ and embed all of these using the \textsc{PoincareEmbeddings} in hyperbolic spaces of dimensions $d \in \mathfrak{D} = [2,5,10,20]$. For a given dimensionality and noise level, we use our method to find correspondences between the noise-less and noisy version of the hierarchy (i.e., $|\mathfrak{P}| \times |\mathfrak{N}|$ matching tasks in total). 

Statistics about all the datasets used in this work are provided in Table~\ref{tab:dataset_details}. Further details about the OAEI datasets can found on the project's website.\footnote{\url{http://oaei.ontologymatching.org/2018/}}

\section{Model Configurations and Hyperparameters}\label{sec:configs}

In Table~\ref{tab:ablation_details}, we provide full configuration details for all the ablated models used in the WordNet \En$\rightarrow$\En self-recovery experiment (results shown in Table~\ref{tab:ablation}). Dashed lines indicate a parameter being the same as in the  \textsc{Full Model}.

\section{A Brief Summary of Theoretical Guarantees for Optimal Transport (Euclidean Case)}\label{sec:appendix_theory_ot_euclidean}

As mentioned in Section~\ref{sec:unsup_matching}, whenever optimal transport is used with the goal of obtaining correspondences, there are various theoretical considerations that become particularly appealing. 

The first of such considerations pertains to the nature of the solution, i.e., the optimal coupling $\pi^*$ which minimizes the cost \eqref{eq:wasserstein_distance}. When the final end goal is to transport points from one space to the other, the best case scenario would be if the optimal $\pi$ happens to be a ``hard'' deterministic mapping. A celebrated result by \citet{brenier1987decomposition} (see also \citep{brenier1991polar}) shows that this indeed the case for the quadratic cost,\footnote{This result holds in more general settings. We refer the reader to \citep{Santambrogio2010Introduction, ambrosio2013users} for further details.} i.e., for the 2-Wasserstein distance. Even when solving the problem approximately with entropic regularization (cf. Eq.~\eqref{eq:entreg_wasserstein}), this result guarantees that the solution found in this way converges to a deterministic mapping as $\varepsilon \rightarrow 0$. 

Now, assuming now that such a map exists, the next aspect we might be interested in is its smoothness. Intuitively, smoothness of this mapping is desirable since it is more likely to lead to robust matchings in the context of correspondences, even if, again, the argument holds asymptotically for the regularized problem. This, clearly, is a very strong property to require. While not even continuity can be guaranteed in general \citep{ambrosio2013users}, again for the quadratic-cost things are simpler: if the source and target densities are smooth and the support of the target distribution satisfies suitable convexity assumptions, the optimal map is guaranteed to be smooth too \citep{ caffarelli1992boundary,caffarelli1992regularity}.

\section{A Brief Summary of Theoretical Guarantees for Optimal Transport (Riemannian Manifold Case)}\label{sec:appendix_theory_ot_manifold}

Extending the problem beyond Euclidean to more general spaces has been one of the central questions theoretical optimal transport research over the past decades \citep{villani2008optimal}. For obvious reasons, here we focus the discussion on results related to hyperbolic spaces, and more generally, to Riemannian manifolds.

Let us first note that Problem \eqref{eq:wasserstein_distance} is well-defined for any complete and separable metric space $\cX$. Since the arclength metric of a Riemannian manifold allows for the direct construction of an accompanying metric space $(\cX, d_{\cX})$, then OT can be defined over those too. However, some of the theoretical results of their Euclidean counterparts do not transfer that easily to the Riemannian case \citep{ambrosio2013users}. Nevertheless, the existence and uniqueness of the optimal transportation plan $\pi^*$, which in addition is induced by a transport map $T$, can be guaranteed with mild regularity conditions on the source distribution $\alpha$. This was first shown in seminal work by \citet{McCann2001Polar}. The result, which acts as an Riemannian analogue of that of Brenier for the Euclidean setting \citep{brenier1987decomposition}, is shown below as presented by \citet{ambrosio2013users}:

\begin{theorem}[McCann, version of \citep{ambrosio2013users}]
    Let $M$ be a smooth, compact Riemannian manifold without boundary and $\alpha \in \cP(\cM)$. Then the following are equivalent:
    \vspace{-0.2cm}
    \begin{enumerate}[wide=0pt, leftmargin=\dimexpr\labelwidth + 2\labelsep\relax]
	    \itemsep1pt
        \item[(i)] $ \forall \beta \in \cP(\cM)$, there exists a unique optimal $\pi \in \Pi(\alpha, \beta)$, and this plan is induced by a map $T$.
        \item[(ii)]  $\alpha$ is regular. 
    \end{enumerate}
    If either (i) or (ii) holds, the optimal $T$ can be written as $x \mapsto \exp_x(-\nabla \phi(x))$ for some c-concave function $\phi:\cM\rightarrow \R$.
\end{theorem}

The question of regularity of the optimal map, on the other hand, is much more delicate now than in the Euclidean case \citep{ambrosio2013users, ma2005regularity, loeper2009regularity}. In addition to the suitable convexity assumptions on the support of the target density, a restrictive structural condition, known as the Ma-Trudinger-Wang (MTW) condition \citep{ma2005regularity}, needs to be imposed on the cost in order to guarantee continuity of the optimal map. Unfortunately for our setting, in the case of Riemannian manifolds the MTW condition for the usual quadratic cost $c = d^2/2$ is so restrictive that it implies that $\cX$ has non-negative sectional curvature \citep{loeper2009regularity}, which rules out hyperbolic spaces. However, a recent sequence of remarkable results \citep{lee2012examples, Li2009SmoothOT} prove that for simple variations of the Riemannian metric $d$ on hyperbolic spaces, smoothness is again guaranteed: 
\begin{theorem}[Lee and Li, \citep{lee2012examples}]\label{thm:mtw_costs}
    Let $d$ be the Riemannian distance function on a manifold of constant sectional curvature $-1$; then the cost functions $-\cosh \circ d$ and $-\log \circ (1+\cosh) \circ d$ satisfy the strong MTW condition, and the cost functions $\pm \log \circ \cosh \circ d$ satisfy the weak MTW condition.
\end{theorem}
Thus, these cost objectives can be used it out hyperbolic optimal transport matching setting with the hopes of obtaining a smoother solution, and therefore a more stable set of correspondences.

\end{document}